\documentclass{article}

    \usepackage[final, nonatbib]{neurips_2024}

\usepackage[utf8]{inputenc} 
\usepackage[T1]{fontenc}    
\usepackage{hyperref}       
\usepackage{url}            
\usepackage{booktabs}       
\usepackage{amsfonts}       
\usepackage{nicefrac}       
\usepackage{microtype}      
\usepackage{xcolor}         
\usepackage{graphicx}
\usepackage{amsmath}
\usepackage{amssymb}
\usepackage{algorithm}
\usepackage{color}
\usepackage{caption,subcaption}
\usepackage{multirow}
\usepackage{blindtext}
\usepackage{hhline}
\usepackage{subcaption}
\usepackage{graphicx}

\newbox{\bigpicturebox}

\def\0{{\mathbf 0}}
\def\1{{\mathbf 1}}

\def\b{{\mathbf b}}

\def\f{{\mathbf f}}
\def\g{{\mathbf g}}

\def\l{{\mathbf l}}

\def\q{{\mathbf q}}

\def\v{{\mathbf v}}

\def\x{{\mathbf x}}
\def\y{{\mathbf y}}
\def\z{{\mathbf z}}

\def\A{{\mathbf A}}
\def\B{{\mathbf B}}
\def\C{{\mathbf C}}
\def\D{{\mathbf D}}

\def\H{{\mathbf H}}
\def\I{{\mathbf I}}

\def\K{{\mathbf K}}
\def\L{{\mathbf L}}
\def\M{{\mathbf M}}

\def\P{{\mathbf P}}
\def\Q{{\mathbf Q}}

\def\V{{\mathbf V}}
\def\W{{\mathbf W}}

\def\ie{{\textit{i.e.}}}
\def\eg{{\textit{e.g.}}}

\def\cE{{\mathcal E}}

\def\cG{{\mathcal G}}

\def\cL{{\mathcal L}}

\def\cN{{\mathcal N}}
\def\cO{{\mathcal O}}

\def\cS{{\mathcal S}}

\def\bphi{{\pmb{\phi}}}

\def\bphi{{\pmb{\phi}}}

\def\bmu{{\boldsymbol \mu}}

\def\bLambda{{\boldsymbol \Lambda}}
\def\bPhi{{\boldsymbol \Phi}}

\def\bTheta{{\boldsymbol \Theta}}

\newcommand{\red}[1] {\textcolor[rgb]{1.0,0.0,0.0}{{#1}}}

\title{Interpretable Lightweight Transformer via Unrolling of Learned Graph Smoothness Priors}

\author{%
  Tam Thuc Do \\
  York University \\
  Toronto, Canada \\
  \texttt{dtamthuc@yorku.ca} \\
  \And
  Parham Eftekhar\\
  York University \\
  Toronto, Canada \\
  \texttt{eftekhar@yorku.ca} \\
  \And
  Seyed Alireza Hosseini\\
  York University \\
  Toronto, Canada \\
  \texttt{ahoseini@yorku.ca} \\
  \AND
  Gene Cheung\\
  York University \\
  Toronto, Canada \\
  \texttt{genec@yorku.ca} \\
  \And
  Philip A. Chou\\
  \tt packet.media\\
  Seattle, USA \\
  \texttt{pachou@ieee.org} \\
}

\begin{document}

\maketitle

\begin{abstract}
We build interpretable and lightweight transformer-like neural networks by unrolling iterative optimization algorithms that minimize graph smoothness priors---the quadratic graph Laplacian regularizer (GLR) and the $\ell_1$-norm graph total variation (GTV)---subject to an interpolation constraint.
The crucial insight is that a normalized signal-dependent graph learning module amounts to a variant of the basic self-attention mechanism in conventional transformers.
Unlike ``black-box'' transformers that require learning of large key, query and value matrices to compute scaled dot products as affinities and subsequent output embeddings, resulting in huge parameter sets, our unrolled networks employ shallow CNNs to learn low-dimensional features per node to establish pairwise Mahalanobis distances and construct sparse similarity graphs.
At each layer, given a learned graph, the target interpolated signal is simply a low-pass filtered output derived from the minimization of an assumed graph smoothness prior, leading to a dramatic reduction in parameter count. 
Experiments for two image interpolation applications verify the restoration performance, parameter efficiency and robustness to covariate shift of our graph-based unrolled networks compared to conventional transformers. 
\end{abstract}

\section{Introduction}
\label{sec:intro}
Focusing on the \textit{self-attention} mechanism \cite{bahdanau14} as the basic building block---where the \textit{affinity} between two input tokens is computed as a transformed dot product---\textit{transformers} \cite{vaswani17attention} learn large parameter sets to achieve state-of-the-art (SOTA) performance in a wide range of signal prediction/classification problems \cite{liu21,dosovitskiy21}, outperforming convolutional neural nets (CNNs) and recurrent neural nets (RNNs). 
However, there are shortcomings: i) lack of mathematical interpretability to characterize general performance\footnote{While works exist to analyze existing transformer architectures \cite{vuckovic20,zhang20,snell21,wei22,kim21}, only \cite{edelman22,li23} characterized the performance of a single self-attention layer and a shallow transformer, respectively. In contrast, we build transformer-like networks by unrolling graph-based algorithms, so that each layer is interpretable by construction.}, ii) requiring substantial training datasets to train sizable parameters \cite{dehghani23}, and iii) fragility to \textit{covariate shift}---when training and testing data have different distributions \cite{zhang24}. 

Orthogonally, \textit{algorithm unrolling} \cite{monga21} implements iterations of a model-based algorithm as a sequence of neural layers to build a feed-forward network, whose parameters can be learned end-to-end via back-propagation from data. 
A classic example is the unrolling of the \textit{iterative soft-thresholding algorithm} (ISTA) in sparse coding into \textit{Learned} ISTA (LISTA) \cite{gregor10}. 
Recently, \cite{yu23nips} showed that by unrolling an iterative algorithm minimizing a sparse rate reduction (SRR) objective, it can lead to a family of ``white-box'' transformer-like deep neural nets that are 100\% mathematically interpretable. 
Inspired by \cite{yu23nips}, in this work we also seek to build white-box transformers via algorithm unrolling, but from a unique \textit{graph signal processing} (GSP) perspective
\cite{shuman13,ortega18ieee,cheung18}. 

Over the last decade and a half, GSP studies spectral analysis and processing of discrete signals on structured data kernels described by finite graphs.
Specifically, by assuming that the sought signal is smooth (low-pass) with respect to (w.r.t.) a particular graph, a plethora of graph-based restoration algorithms can be designed for practical applications, including image denoising \cite{pang17}, JPEG dequantization \cite{liu17}, interpolation \cite{chen24}, 3D point cloud denoising \cite{zeng19,dinesh20} and super-resolution \cite{dinesh22}. 
At the heart of GSP is the construction of a \textit{similarity graph} that captures pairwise similarities between signal samples on two connected nodes. 
We first demonstrate that \textit{signal-dependent similarity graph learning with normalization is akin to affinity computation in the self-attention mechanism}.
Thus, \textit{our first contribution is to show that unrolling of a graph-based iterative algorithm\footnote{While there exist works on unrolling of graph-based algorithms \cite{yang21g}, they typically assume a \textit{fixed} graph and optimize other parameters. In contrast, the key point in our work is that the normalized graph learning module is akin to the basic self-attention mechanism in conventional transformers. } with normalized graph learning results in an interpretable transformer-like feed-forward network}.

Second, computation of a positive weight $w_{i,j} = \exp(-d(i,j))$ of an edge $(i,j)$ connecting two graph nodes $i$ and $j$ often employs \textit{Mahalanobis distance} $d(i,j) = (\f_i - \f_j)^\top \M (\f_i - \f_j)$ between representative (\eg, CNN-computed) \textit{feature vectors} $\f_i$ and $\f_j$, where $\f_i, \f_j \in \mathbb{R}^D$ reside in low-dimensional space \cite{hu20,yang22}. 
Hence, unlike a conventional transformer that requires large key and query matrices, $\K$ and $\Q$, to compute transformed dot products, the similarity graph learning module can be more parameter-efficient.
Moreover, by adding a graph smoothness prior such as \textit{graph Laplacian regularizer} (GLR) \cite{ortega18ieee,pang17} or \textit{graph total variation} (GTV) \cite{elmoataz08,couprie13,berger17} in the optimization objective, once a graph $\cG$ is learned, the target signal is simply computed as the low-pass filtered output derived from the minimization of the assumed graph smoothness prior.
Thus, a large value matrix $\V$ to compute output embeddings typical in a transformer is also not needed.
\textit{Our second contribution is to demonstrate that a \textbf{lightweight} transformer with fewer parameters can be built via unrolling of a graph-based restoration algorithm with a chosen graph signal smoothness prior}. 

Specifically, focusing on the signal interpolation problem, we first derive linear-time graph-based algorithms by minimizing GLR or GTV, via \textit{conjugate gradient} (CG) \cite{shewchuk94} or a modern adaptation of \textit{alternative method of multipliers} (ADMM) for \textit{sparse linear programming} (SLP) \cite{wang17}, respectively.
In each iteration, given a learned graph, each algorithm deploys a low-pass graph filter to interpolate the up-sampled observation vector into the target signal.
We intersperse unrolled algorithm iterations with graph learning modules into a compact and interpretable neural network.
We demonstrate its restoration performance, parameter efficiency ($3\%$ of SOTA's parameters in one case), and robustness to covariate shift  for two practical applications: image demosaicking, and image interpolation. 


\vspace{0.1in}
\noindent
\textbf{Notation:}
Vectors and matrices are written in bold lowercase and uppercase letters, respectively.
The $(i,j)$ element and the $j$-th column of a matrix $\mathbf{A}$ are denoted by $A_{i,j}$ and $\mathbf{a}_{j}$, respectively.
The $i$-th element in the vector $\mathbf{a}$ is denoted by $a_{i}$.
The square identity matrix of rank $N$ is denoted by $\mathbf{I}_N$, the $M$-by-$N$ zero matrix is denoted by $\mathbf{0}_{M,N}$, and the vector of all ones / zeros of length $N$ is denoted by $\mathbf{1}_N$ / $\mathbf{0}_N$, respectively.
Operator $\|\cdot\|_p$ denotes the $\ell$-$p$ norm. 


\vspace{-0.05in}
\section{Preliminaries}
\label{sec:prelim}
\subsection{GSP Definitions}
\label{subsec:defn}

A graph $\cG(\cN,\cE,\W)$ is defined by a node set $\cN = \{1, \ldots, N\}$ and an edge set $\cE$ of size $|\cE| = M$, where $(i,j) \in \cE$ means nodes $i,j \in \cN$ are connected with weight $w_{i,j} = W_{i,j} \in \mathbb{R}$.
In this paper, we consider only \textit{positive} graphs $\cG$ with no self-loops, \ie, $w_{i,j} \geq 0, \forall i,j$, and $w_{i,i} = 0, \forall i$.  
We assume edges are undirected, and thus \textit{adjacency matrix} $\W \in \mathbb{R}^{N \times N}$ is symmetric. 
The \textit{combinatorial graph Laplacian matrix} is defined as $\L \triangleq \D - \W \in \mathbb{R}^{N \times N}$, where $\D \triangleq \text{diag}(\W \1_N)$ is the \textit{degree matrix}, and $\text{diag}(\v)$ returns a diagonal matrix with $\v$ along its diagonal.
$\L$ for a positive graph $\cG$ is provably \textit{positive semi-definite} (PSD), \ie, all its eigenvalues $\lambda_i$'s are non-negative \cite{cheung18}.

We define also the \textit{incidence matrix} $\C = \mathbb{R}^{M \times N}$: each $k$-th row of $\C$ corresponds to the $k$-th edge $(i,j) \in \cE$, where $C_{k,i} = w_{i,j}$, $C_{k,j} = -w_{i,j}$, and $C_{k,l} = 0, ~\forall l \neq i,j$. 
Since our assumed graph $\cG$ is undirected, the polarities of $C_{k,i}$ and $C_{k,j}$ are arbitrary, as long as they are opposite.


\subsection{Graph Laplacian Regularizer}
\label{subsec:GLR}

Given a positive \textit{connected} graph $\cG$ with $N$ nodes and $M$ edges, we first define smoothness of a signal $\x \in \mathbb{R}^N$ w.r.t. $\cG$ using the \textit{graph Laplacian regularizer} (GLR) \cite{ortega18ieee,pang17} as
\begin{align}
\|\x\|_{\cG,2} = \x^\top \L \x = \sum_{(i,j) \in \cE} w_{i,j} (x_i - x_j)^2 
\label{eq:GLR}
\end{align}
where $\L$ is a combinatorial Laplacian matrix specifying graph $\cG$. 
GLR \eqref{eq:GLR} is non-negative for a positive graph, and thus is suitable as a signal prior for minimization problems \cite{pang17,liu17}. 

\subsection{Graph Total Variation}
\label{subsec:GTV}

Instead of GLR \eqref{eq:GLR}, we can alternatively define graph signal smoothness using \textit{graph total variation} (GTV) \cite{elmoataz08,couprie13,berger17} $\|\x\|_{\cG,1}$ for signal $\x \in \mathbb{R}^N$ as
\begin{align}
\|\x\|_{\cG,1} = \|\C \x\|_1 \stackrel{(a)}{=} \sum_{(i,j) \in \cE} w_{i,j} | x_i - x_j |
\end{align}
where $(a)$ is true since $\cG$ is positive. 
GTV is also non-negative for positive $\cG$, and has been used as a signal prior for restoration problems such as image deblurring \cite{bai19}.


\vspace{-0.05in}
\section{Problem Formulation \& Optimization using GLR}
\label{sec:formulate}
\subsection{Problem Formulation}

We first assume a positive, \textit{sparse} and \textit{connected} graph $\cG$ with $N$ nodes and $M$ edges specified by graph Laplacian matrix $\L$.
By sparse, we mean that $M$ is $\cO(N)$ and not $\cO(N^2)$. 
By connected, we mean that any node $j$ can be traversed from any other node $i$. 
Given $\cG$, we first derive a \textit{linear-time} iterative algorithm to interpolate signal $\x$ by minimizing GLR given observed samples $\y$. 
In Section\;\ref{sec:formulate2}, we derive an algorithm by minimizing GTV instead given $\y$. 
In Section\;\ref{sec:learn}, we unroll iterations of one of two derived algorithms into neural layers, together with strategically inserted graph learning modules, to construct graph-based lightweight transformer-like neural nets.

We first employ GLR \cite{pang17} as the objective to reconstruct $\x \in \mathbb{R}^N$ given partial observation $\y \in \mathbb{R}^K$, where $K < N$. 
Denote by $\H \in \{0,1\}^{K \times N}$ a \textit{sampling matrix} defined as
\begin{align}
H_{i,j} = \left\{ \begin{array}{ll} 
1 & \mbox{if node $j$ is the $i$-th sample} \\
0 & \mbox{o.w.}
\end{array} \right. 
\end{align}
that picks out $K$ samples from signal $\x$. 
The optimization is thus 
\begin{align}
\min_{\x} ~ \x^\top \L \x, 
~~~~~~ \mbox{s.t.}~~ \H \x = \y 
\label{eq:GLR_inter}
\end{align}
where $\L \in \mathbb{R}^{N \times N}$ is a graph Laplacian matrix corresponding to a positive graph $\cG$ \cite{cheung18}. 
PSD $\L$ implies that $\x^\top \L \x \geq 0, \forall \x$, and thus \eqref{eq:GLR_inter} has a convex objective with a linear interpolation constraint.

\subsection{Optimization}

We solve \eqref{eq:GLR_inter} via a standard Lagrangian approach \cite{boyd04} and write its corresponding unconstrained Lagrangian function $f(\x, \bmu)$ as
\begin{align}
f(\x,\bmu) = \x^\top \L \x + \bmu^\top (\H \x - \y) 
\label{eq:GLR_lagrange}
\end{align}
where $\bmu \in \mathbb{R}^K$ is the Lagrange multiplier vector. 
To minimize $f(\x, \bmu)$ in \eqref{eq:GLR_lagrange}, we take the derivative w.r.t. $\x$ and $\bmu$ separately and set them to zero, resulting in the following linear system:
\begin{align}
\underbrace{\left[ \begin{array}{cc}
2 \L & \H^\top \\
\H & \0_{K,K} 
\end{array}
\right]}_{\P}
\left[ \begin{array}{c}
\x \\
\bmu 
\end{array} \right] &= 
\left[ \begin{array}{c} 
\0_{N,N} \\
\y
\end{array} \right] .
\label{eq:GLR_sol}
\end{align}

Given that the underlying graph $\cG$ is positive and connected, coefficient matrix $\P$ is provably full-rank and thus invertible (see Appendix\;\ref{append:GLR} for a proof).
Hence, \eqref{eq:GLR_sol} has a unique solution $\x^*$. 

Suppose we index the sampled nodes $\cS$ in $\x$ before the non-sampled nodes $\bar{\cS}$, \ie, $\x = [\x_{\cS}; \x_{\bar{\cS}}]$.
Then $\H = [\I_{K} ~~\0_{K,N-K}]$, and the second block row in \eqref{eq:GLR_sol} implies $\x_{\cS} = \y$. 
Suppose we write $\L = [\L_{\cS,\cS} ~~ \L_{\cS,\bar{\cS}}; \L_{\bar{\cS},\cS} ~~ \L_{\bar{\cS},\bar{\cS}}]$ in blocks also.
For the first block row in \eqref{eq:GLR_sol}, consider only the non-sampled rows:
\begin{align}
\left( \left[ 2\L ~~ \H^\top \right] \left[ \begin{array}{c}
\x \\
\bmu 
\end{array} \right] \right)_{\bar{\cS}} = 2 (\L_{\bar{\cS},\cS} \, \x_{\cS} + \L_{\bar{\cS},\bar{\cS}} \, \x_{\bar{\cS}} ) = \0_{N-K}
\end{align}
where $\left( \H^\top \bmu \right)_{\bar{\cS}} = \0_{N-K}$ since the non-sampled rows of $\H^\top$ (the non-sampled columns of $\H$) are zeros. 
Thus, $\x_{\bar{\cS}}$ can be computed via the following system of linear equations:
\begin{align}
\L_{\bar{\cS},\bar{\cS}} \,  \x_{\bar{\cS}} = - \L_{\bar{\cS},\cS} \, \y
\label{eq:GLR_sol2}
\end{align}
where $\L_{\bar{\cS},\bar{\cS}}$ is a symmetric, sparse, and provably \textit{positive definite} (PD) matrix (see Appendix\;\ref{append:PD_L} for a proof). 
Thus, there exists a unique solution $\x_{\bar{\cS}}$ in \eqref{eq:GLR_sol2}.

\textbf{Complexity}: 
For notation simplicity, let $\cL = \L_{\bar{\cS},\bar{\cS}}$.
Linear system \eqref{eq:GLR_sol2} can be solved efficiently using \textit{conjugate gradient} (CG), an iterative descent algorithm with complexity $\mathcal{O}(\text{nnz}(\boldsymbol{\cL}) \sqrt{\kappa(\boldsymbol{\cL})}/\log(\epsilon))$, where $\text{nnz}(\boldsymbol{\cL})$ is the number of non-zero entries in matrix $\boldsymbol{\cL}$, $\kappa(\cL) = \frac{\lambda_{\max}(\boldsymbol{\cL})}{\lambda_{\min}(\boldsymbol{\cL})}$ is the \textit{condition number} of $\boldsymbol{\cL}$,  $\lambda_{\max}(\boldsymbol{\cL})$ and $\lambda_{\min}(\boldsymbol{\cL})$ are the respective largest and smallest eigenvalues of $\boldsymbol{\cL}$, and $\epsilon$ is the convergence threshold of the gradient search \cite{shewchuk94}. 
Because $\L$ is sparse by graph construction ($\cO(N)$ edges), $\cL = \L_{\bar{\cS},\bar{\cS}}$ is also sparse, \ie, $\text{nnz}(\cL) = \cO(N)$.  
Assuming $\kappa(\cL)$ can be reasonably lower-bounded for PD $\cL$ and $\epsilon$ is reasonably chosen, the complexity of solving \eqref{eq:GLR_sol2} using CG is $\cO(N)$. 

\textbf{Interpretation}: 
To elicit a signal filtering interpretation from \eqref{eq:GLR_sol}, we assume for now that $\L$ is PD\footnote{A combinatorial graph Laplacian matrix for a positive graph with at least one additional positive self-loop is provably PD; see proof in Appendix\;\ref{append:PD_L}.} and thus invertible.
Recall that the block matrix inversion formula \cite{golub12} is 
\begin{align}
\P^{-1} &= \left( \left[ \begin{array}{cc}
\A & \B \\
\C & \D \end{array} \right] \right)^{-1} = 
\left[ \begin{array}{cc}
\A^{-1} + \A^{-1} \B (\P/\A)  \C \A^{-1} & -\A^{-1} \B (\P/\A) \\
-(\P/\A) \C \A^{-1} & (\P/\A)
\end{array} \right]
\end{align}
where $\P/\A = (\D - \C \A^{-1} \B)^{-1}$ is the \textit{Schur complement} of block $\A$ of matrix $\P$. 
Solution $\x^*$ can thus be computed as:
\begin{align}
\x^* &= \L^{-1} \H^\top \left( \H \L^{-1} \H^\top \right)^{-1} \y 
\nonumber \\
&= \L^{-1} \H^\top \left( (\L^{-1})_\cS \right)^{-1} \y = \L^{-1} \H^\top \L_\cS^\# \y
\label{eq:GLR_filter}
\end{align}
where $\L_{\cS}^\# \triangleq \left( (\L^{-1})_\cS \right)^{-1}$ and $(\L^{-1})_\cS$ denotes the rows and columns of $\L^{-1}$ corresponding to the sampled nodes.
$\L_{\cS}^\#$ is a high-pass filter similar to $\L_{\cS}$.
Thus, we can interpret $\x^*$ as a \textit{low-pass filtered output of up-sampled $\H^\top \L_\cS^\# \y$}---with low-pass filter response $r(\bLambda) = \bLambda^{-1}$ where $\L = \V \bLambda \V^{\top}$, $\bLambda = \text{diag}([\lambda_1, \ldots, \lambda_N])$, is eigen-decomposible with frequencies $\lambda_k$'s and Fourier modes $\v_k$'s. 

\vspace{-0.05in}
\section{Problem Formulation \& Optimization using GTV}
\label{sec:formulate2}
\subsection{Problem Formulation}

Given a positive connected graph $\cG$ specified by incidence matrix $\C \in \mathbb{R}^{M \times N}$ and partial observation $\y$, we now employ instead GTV as the objective to interpolate target signal $\x$, resulting in
\begin{align}
\min_{\x} \; \|\C \x\|_1, 
~~~~~~ \mbox{s.t.}~~ \H \x = \y ,
\label{eq:GTV_inter}
\end{align}
\eqref{eq:GTV_inter} is a \textit{linear program} (LP), since both the objective and the lone constraint are linear. 
Thus, while minimizing GLR leads to a \textit{linear system} \eqref{eq:GLR_sol}, minimizing GTV leads to a \textit{linear program} \eqref{eq:GTV_inter}.

\subsubsection{LP in Standard Form}

First, we rewrite LP \eqref{eq:GTV_inter} in standard form as follows.
Define \textit{upper-bound variable} $\z \in \mathbb{R}^M$ with a pair of linear constraints $\z \geq \pm \C \x$. 
This enables a linear objective $\1_M^\top \z$ for a minimization problem (thus ensuring the upper bound is tight), \ie, $\z = \|\C \x\|_1$. 
Second, we introduce non-negative \textit{slack variables} $\q_1, \q_2 \in \mathbb{R}^M$ to convert inequality constraints $\z \geq \pm \C \x$ to equality constraints $\z = \C \x + \q_1$ and $\z = -\C \x + \q_2$. 
Thus, LP \eqref{eq:GTV_inter} can be rewritten as
\begin{align}
\min_{\z,\x,\q} \1_M^\top \z, ~~~ \mbox{s.t.}~
\underbrace{\left[ \begin{array}{ccc}
\I_M & -\C & -(\I_M ~~ \0_{M,M}) \\
\I_M &  \C & -(\0_{M,M} ~~ \I_M) \\
\0_{K,M} & \H & \0_{K,2M}
\end{array} \right]}_{\A}
\left[ \begin{array}{c}
\z \\
\x \\
\q
\end{array} \right] = 
\underbrace{\left[ \begin{array}{c}
\0_M \\
\0_M \\
\y 
\end{array} \right]}_{\b}, 
~~~~~ \q \geq \0_{2M}
\label{eq:LP_standard}
\end{align}
where $\q = [\q_1; \q_2] \in \mathbb{R}^{2M}$.

\subsection{Optimization Algorithm}
\label{subsec:ADMM_opt}

Because coefficient matrix $\A$ is sparse, \eqref{eq:LP_standard} is a \textit{sparse linear program} (SLP). 
We solve SLP \eqref{eq:LP_standard} efficiently by adopting an ADMM approach for SLP in \cite{wang17}. 
We first define a convex
but non-differentiable (non-smooth) \textit{indicator function}:
\begin{align}
g(\q) = \left\{ \begin{array}{ll}
0 & \mbox{if}~ q_j \geq 0, ~~ \forall j \\
\infty & \mbox{o.w.}
\end{array} \right. .
\end{align}

We next introduce \textit{auxiliary variable} $\tilde{\q} \in \mathbb{R}^{2M}$ and equality constraint $\tilde{\q} = \q$. 
We now rewrite \eqref{eq:LP_standard} with a single equality constraint as
\begin{align}
\min_{\z,\x,\q,\tilde{\q}} \1_M^\top \z + g(\tilde{\q}), ~~~
\mbox{s.t.}~ 
\underbrace{\left[ \begin{array}{c}
\A \\
\0_{2M,M+N} ~ \I_{2M} \end{array} \right]}_{\B} 
\left[ \begin{array}{c}
\z \\
\x \\
\q 
\end{array} \right]
= \left[ \begin{array}{c}
\b \\
\tilde{\q}
\end{array} \right] .
\label{eq:opt_eq}
\end{align}
We can now rewrite \eqref{eq:opt_eq} into an unconstrained version using the augmented Lagrangian method as
\begin{align}
\min_{\z, \x, \q, \tilde{\q}} \1_M^\top \z + g(\tilde{\q})
+ \bmu^\top \left( \B \left[ \begin{array}{c}
\z \\
\x \\
\q
\end{array} \right] - \left[ \begin{array}{c}
\b \\
\tilde{\q}
\end{array} \right] \right)
+ \frac{\gamma}{2} \left\| \B \left[ \begin{array}{c}
\z \\
\x \\
\q
\end{array} \right] - \left[ \begin{array}{c}
\b \\
\tilde{\q}
\end{array} \right] \right\|^2_2
\label{eq:obj_ADMM}
\end{align}
where $\bmu \in \mathbb{R}^{4M+K}$ is a Lagrange multiplier vector, and $\gamma > 0$ is a scalar parameter. 
In the sequel, we write $\bmu = [\bmu_a; \bmu_b; \bmu_c; \bmu_d; \bmu_e]$, where $\bmu_a, \bmu_b, \bmu_d, \bmu_e \in \mathbb{R}^M$ and $\bmu_c \in \mathbb{R}^K$.

\subsubsection{Optimizing Main Variables}

As typically done in ADMM, we minimize the unconstrained objective \eqref{eq:obj_ADMM} alternately as follows.
At iteration $t$, when $\tilde{\mathbf{q}}^t$ and $\bmu^t$ are fixed, the optimization for $\z^{t+1}$, $\x^{t+1}$ and $\q^{t+1}$ becomes
\begin{align}
\min_{\z, \x, \q} \1_M^\top \z + (\bmu^t)^\top \left( \B \left[ \begin{array}{c}
\z \\
\x \\
\q
\end{array} \right] - \left[ \begin{array}{c}
\b \\
\tilde{\q}
\end{array} \right] \right)
+ \frac{\gamma}{2} \left\| \B \left[ \begin{array}{c}
\z \\
\x \\
\q
\end{array} \right] - \left[ \begin{array}{c}
\b \\
\tilde{\q}
\end{array} \right] \right\|^2_2 .
\label{eq:obj_ADMM1}
\end{align}
The solution to this convex and smooth quadratic optimization is a system of linear equations,

\vspace{-0.1in}
\begin{small}
\begin{align}
\z^{t+1} &= -\frac{1}{\gamma} \1_M - \frac{1}{2\gamma} \left( \bmu_a^t + \bmu_b^t + \bmu_d^t + \bmu_e^t \right) + \frac{1}{2} (\tilde{\q}_1^t + \tilde{\q}_2^t)  
\label{eq:sol_ADMM1a} \\
(\C^\top \C + \H^\top \H) \x^{t+1} &= 
 \frac{1}{2\gamma} \C^\top \left( \bmu_a^t - \bmu_b^t + \bmu_d^t - \bmu_e^t \right) - \frac{1}{\gamma} \H^\top \bmu_c^t - \frac{1}{2} \C^\top (\tilde{\q}_1^t - \tilde{\q}_2^t) + \H^\top \y   
\label{eq:sol_ADMM1b} \\
\q_1^t &= \frac{1}{2} \left( \z^{t+1} - \C \x^{t+1} \right) + \frac{1}{2\gamma}(\bmu_a^t - \bmu_d^t + \gamma \tilde{\q}_1^t)
\nonumber \\
\q_2^t &= \frac{1}{2} \left( \z^{t+1} + \C \x^{t+1} \right) + \frac{1}{2\gamma}(\bmu_b^t - \bmu_e^t + \gamma \tilde{\q}_2^t)
\label{eq:sol_ADMM1c}
\end{align}
\end{small}\noindent
See the Appendix\;\ref{append:sol_ADMM1} for a derivation.

Linear system \eqref{eq:sol_ADMM1b} is solvable if the coefficient matrix $\boldsymbol{\cL} \triangleq \L + \H^\top \H$, where $\L \triangleq \C^\top \C$ is a PSD graph Laplacian for a positive graph, is invertible. 
See Appendix\;\ref{append:invertibility} for a proof that $\cL$ is PD and thus invertible .

\textbf{Complexity}:
Linear system \eqref{eq:sol_ADMM1b} again can be solved efficiently using CG with complexity $\mathcal{O}(\text{nnz}(\boldsymbol{\cL}) \sqrt{\kappa(\boldsymbol{\cL})}/\log(\epsilon))$ \cite{shewchuk94}. 
Because $\C$ is sparse by graph construction ($\cO(N)$ edges) and $\H$ is sparse by definition, $\cL$ is also sparse, \ie, $\text{nnz}(\cL) = \cO(N)$.  
Thus, assuming $\kappa(\cL)$ is also upper-bounded and $\epsilon$ is reasonably chosen, the complexity of solving \eqref{eq:sol_ADMM1b} using CG is $\cO(N)$. 

\textbf{Interpretation}:
\eqref{eq:sol_ADMM1b} can be interpreted as follows.
$\H^\top \y$ is the \textit{up-sampled version} of observation $\y$.
$\cL = \V \bLambda \V^\top$, $\Lambda = \text{diag}([\lambda_1, \ldots, \lambda_N])$, is an eigen-decomposible \textit{generalized Laplacian matrix}---Laplacian matrix $\C^\top \C$ plus self-loops of weight $1$ at sampled nodes due to diagonal matrix $\H^\top \H$, and like a graph Laplacian $\L$ without self-loops, can be interpreted as a high-pass spectral filter \cite{davies00}. 
$\cL^{-1} = \V \bLambda^{-1} \V^\top$ is thus 
a \textit{low-pass} spectral filter with frequency response $r(\bLambda) = \bLambda^{-1}$ to interpolate output $\cL^{-1} \H^\top \y$. 
We interpret the remaining terms on the right-hand side of \eqref{eq:sol_ADMM1b} as bias.

\subsubsection{Optimizing Auxiliary Variable}

Fixing $\z^{t+1}$, $\x^{t+1}$ and $\q^{t+1}$, the optimization for $\tilde{\q}^{t+1}$ for \eqref{eq:obj_ADMM} simplifies to 
\begin{align}
\min_{\tilde{\q}} g(\tilde{\q})
+ (\bmu^t_{d})^\top \left( \q^{t+1} - \tilde{\q} \right) 
+ \frac{\gamma}{2} \left\| \q^{t+1} - \tilde{\q} 
\right\|^2_2 .
\label{eq:obj_ADMM2}
\end{align}
The solution for optimal $\tilde{\q}^{t+1}$ is term-by-term thresholding:
\begin{align}
\tilde{q}_i^{t+1} &= \left\{ \begin{array}{ll}
q_i^{t+1} + \frac{1}{\gamma} \mu^t_{d,i} & \mbox{if}~~ q_i^{t+1} + \frac{1}{\gamma} \mu_{d,i}^t \geq 0 \\  
0 & \mbox{o.w.}
\end{array} \right. , \forall i. 
\label{eq:sol_ADMM2}
\end{align}
See Appendix\;\ref{append:sol_ADMM2} for a derivation.

\subsubsection{Updating Lagrange Multiplier}

The Lagrange multiplier $\bmu^{t+1}$ can be updated in the usual manner in an ADMM framework \cite{boyd11}:
\begin{align}
\bmu^{t+1} &= \bmu^t + \gamma  \left( \B \left[ \begin{array}{c} 
\z \\
\x \\
\q 
\end{array} \right] - 
\left[ \begin{array}{c}
\b \\
\tilde{\q} \end{array} \right] \right) .
\label{eq:sol_ADMM3}
\end{align}

\textbf{Algorithm Complexity}: 
Given that the number of iterations until ADMM convergence is not a function of input size, it is $\cO(1)$. 
The most time-consuming step in each ADMM iteration is the solving of linear system \eqref{eq:sol_ADMM1b} via CG in $\cO(N)$.
Thus, we conclude that solving SLP \eqref{eq:LP_standard} using the aforementioned ADMM algorithm is $\cO(N)$. 

\textbf{Algorithm Comparison}: Comparing the CG algorithm used to solve linear system \eqref{eq:GLR_sol2} and the ADMM algorithm developed to solve SLP \eqref{eq:LP_standard}, we first observe that, given a similarity graph $\cG$ specified by Laplacian or incidence matrix, $\L$ or $\C$, both algorithms compute the interpolated signal $\x^*$ as a low-pass filtered output of the up-sampled input $\H^\top \y$ in \eqref{eq:GLR_filter} and \eqref{eq:sol_ADMM1b}, respectively.
This is intuitive, given the assumed graph smoothness priors, GLR and GTV.
We see also that the ADMM algorithm is more intricate: in each iteration, the main variables are computed using CG, while the auxiliary variable is updated via ReLU-like thresholding.  
As a result, the ADMM algorithm is more amenable to deep algorithm unrolling with better performance in general (see Section\;\ref{sec:results} for details).

\vspace{-0.05in}
\section{Graph Learning \& Algorithm Unrolling}
\label{sec:learn}

We now discuss how a similarity graph $\cG$ can be learned from data, specified by graph Laplacian $\L$ for GLR minimization \eqref{eq:GLR_inter} or incidence matrix $\C$ for GTV minimization \eqref{eq:GTV_inter}, so that the two proposed graph-based interpolations can take place.
Moreover, we show how a normalized graph learning module\footnote{While estimating a precision (graph Laplacian) matrix from an empirical covariance matrix computed from data is another graph learning approach \cite{egilmez17,dong19,bagheri24}, we pursue a feature-based approach here \cite{hu20,yang22}.} performs comparable operations to the self-attention mechanism in conventional transformers. 
Thus, unrolling sequential pairs of graph-based iterative algorithm and graph learning module back-to-back leads to an interpretable ``white-box'' transformer-like neural net.

\subsection{Self-Attention Operator in Transformer}

We first review the self-attention operator in a conventional transformer architecture, defined using a transformed dot product and a softmax operation \cite{bahdanau14}.
Specifically, first denote by $\x_i \in \mathbb{R}^E$ an \textit{embedding} for token $i$ of $N$ tokens.
\textit{Affinity} $e(i,j)$ between tokens $i$ and $j$ is defined as the dot product between linear-transformed embeddings $\K \x_i$ and $\Q \x_j$, where $\Q, \K \in \mathbb{R}^{E \times E}$ are the \textit{query} and \textit{key} matrices, respectively.
Using softmax, a non-linear function that maps a vector of real numbers to a vector of positive numbers that sum to $1$, \textit{attention weight} $a_{i,j}$ is computed as 
\begin{align}
a_{i,j} = \frac{\exp (e(i,j))}{\sum_{l=1}^N \exp (e(i,l))}, ~~~~~~ 
e(i,j) = (\Q \x_j)^\top (\K \x_i) .
\label{eq:SA_coeff}
\end{align}

Given self-attention weights $a_{i,j}$, output embedding $\y_i$ for token $i$ is computed as
\begin{align}
\y_i = \sum_{l=1}^N a_{i,l} \x_l \V 
\label{eq:SA_operator}
\end{align}
where $\V \in \mathbb{R}^{E \times E}$ is a \textit{value} matrix.
``Self-attention'' here means that input embeddings are weighted to compute output embeddings.
A transformer is thus a sequence of embedding-to-embedding mappings via different learned self-attention operations defined by $\Q$, $\K$ and $\V$ matrices.
\textit{Multi-head} attention is possible when multiple query and key matrices $\Q^{(m)}$ and $\K^{(m)}$ are used to compute different attention weights $a^{(m)}_{i,j}$'s for the same input embeddings $\x_i$ and $\x_j$, and the output embedding $\y_i$ is computed using an average of these multi-head attention weights $a^{(m)}_{i,l}$'s.


\subsection{Computation of Graph Edge Weights}
\label{subsec:edgeWeight_compute}

Consider now how edge weights $w_{i,j}$'s can be computed from data to specify a finite graph $\cG$ \cite{hu20,yang22}.
A low-dimensional \textit{feature vector} $\f_i \in \mathbb{R}^D$ can be computed for each node $i$ from embedding $\x_i \in \mathbb{R}^E$ via some (possibly non-linear) function $\f_i = F(\x_i)$, where typically $D \ll E$. 
Edge weight $w_{i,j}$ between nodes $i$ and $j$ in a graph $\cG$ can then be computed as
\begin{align}
w_{i,j} &= \exp \left( - d(i, j) \right), ~~~~~~
d(i, j) = (\f_i - \f_j)^\top \M (\f_i - \f_j)
\label{eq:edgeWeight}
\end{align}
where $d(i,j)$ is the squared \textit{Mahalanobis distance} given PSD \textit{metric matrix} $\M$ that quantifies the difference between nodes $i$ and $j$.
$M$ edge weights $\{w_{i,j}\}$ compose a graph $\cG$, specified by the Laplacian matrix $\L$ for GLR minimization \eqref{eq:GLR_inter} and the incidence matrix $\C^{M \times N}$ for GTV minimization \eqref{eq:GTV_inter}. 
Because $w_{i,j} \geq 0, \forall i,j$, constructed graph $\cG$ is positive.

As a concrete example, consider \textit{bilateral filter} (BF) weights commonly used in image filtering \cite{tomasi98}, where feature $\f_i$ contains the 2D grid location $\l_i$ and color intensity $p_i$ of pixel $i$, and metric $\M = \text{diag}([1/\sigma_d^2; 1/\sigma_r^2])$ is a diagonal matrix with weights to specify the relative strength of the \textit{domain} and \textit{range} filters in BF. 
Because BF uses input pixel intensities $p_l$'s to compute weighted output pixel intensities $p_i$'s, BF is \textit{signal-dependent}, similar to self-attention weights in transformers.

Edge weights are often first \textit{normalized} before being used for filtering. 

\textbf{Normalization}: 
For normalization, the symmetric \textit{normalized graph Laplacian} $\L_{n}$ is defined as $\L_{n} \triangleq \D^{-1/2} \L \D^{-1/2}$, so that the diagonal entries of $\L_n$ are all ones (assuming $\cG$ is connected and positive) \cite{ortega18ieee}. 
We assume normalized $\L_n$ is used for Laplacian $\L$ in GLR minimization in \eqref{eq:GLR_inter}.

Alternatively, the asymmetric \textit{random walk graph Laplacian} $\L_{rw}$ is defined as $\L_{rw} \triangleq \D^{-1} \L$, so that the sum of each row of $\L_{rw}$ equals to zero \cite{ortega18ieee}.
Interpreting $\L_{rw}$ as a Laplacian matrix to a \textit{directed} graph, the weight sum of edges leaving each node $i$ is one, \ie, $\sum_{l|(i,l)\in \cE} \bar{w}_{i,l} = 1, \forall i$. 
To accomplish this, undirected edges weights $\{w_{i,j}\}$ are normalized to $\{\bar{w}_{i,j}\}$ via
\begin{align}
\bar{w}_{i,j} = \frac{\exp(-d(i,j))}{\sum_{l | (i,l) \in \cE} \exp(-d(i,l))} .
\label{eq:rwEdgeWeight}
\end{align}
For GTV minimization in \eqref{eq:GTV_inter}, we normalize edge weights in incidence matrix $\C$ instead using \eqref{eq:rwEdgeWeight}.
This results in normalized $\bar{\C} \in \mathbb{R}^{2M \times N}$ for a \textit{directed} graph with $2M$ directed edges.
Subsequently, we define symmetric graph Laplacian $\bar{\L} = \bar{\C}^\top \bar{\C}$ and generalized graph Laplacian $\bar{\cL} = \bar{\L} + \H^\top \H$. 
Note that $\|\bar{\C} \1\|_1 = \sum_{l=1}^N \bar{w}_{i,l} \, |1 - 1| = 0$ after normalization, as expected for a total variation term on a constant signal $\1$. 
Further, note that while $\bar{\C}$ is an incidence matrix for a directed graph with $2M$ edges, $\bar{\L}$ is a graph Laplacian for an undirected graph with $M$ edges.
See Fig.\;\ref{fig:incidence_ex} in Appendix\;\ref{append:normalize} for an example of incidence matrix $\C$, normalized incidence matrix $\bar{\C}$, and graph Laplacian matrix $\bar{\L}$. 

\textbf{Comparison to Self-Attention Operator}: 
We see how the definitions of edge weights \eqref{eq:edgeWeight} and normalization \eqref{eq:rwEdgeWeight} are similar to attention weights in \eqref{eq:SA_coeff}.  
Specifically, \textit{interpreting the negative squared Mahalanobis distance $-d(i,j)$ as affinity $e(i,j)$, normalized edge weights $\bar{w}_{i,j}$ in \eqref{eq:edgeWeight} are essentially the same as attention weights $a_{i,j}$ in \eqref{eq:SA_coeff}}. 
There are subtle but important differences: i) how non-negative Mahalanobis distance $d(i,j)$ is computed in \eqref{eq:edgeWeight} using features $\f_i = F(\x_i)$ and metric $\M$ versus how real-valued affinity is computed via a transformed dot product in \eqref{eq:SA_coeff}, and ii) how the normalization term is computed in a one-hop neighborhood from node $i$ in \eqref{eq:rwEdgeWeight} versus how it is computed using all $N$ tokens in \eqref{eq:SA_coeff}.
The first difference conceptually means that edge weight based on Mahalanobis distance $d(i,j)$ is symmetric (\ie, $\bar{w}_{i,j} = \bar{w}_{j,i}$), while attention weight $a_{i,j}$ is not.  
Both differences have crucial complexity implications, which we will revisit in the sequel.

Further, we note that, given a graph $\cG$, the interpolated signal $\x^*$ is computed simply as a low-pass filtered output of the up-sampled input observation $\H^\top \y$ via \eqref{eq:GLR_filter} or \eqref{eq:sol_ADMM1b}, depending on the assumed graph smoothness prior,  GLR or GTV, while the output embedding $\y_i$ in a conventional transformer requires value matrix $\V$ in \eqref{eq:SA_operator}.
This also has a complexity implication.

\subsection{Deep Algorithm Unrolling}

\begin{figure}[tb]
\centering
\includegraphics[width = 0.8\linewidth]{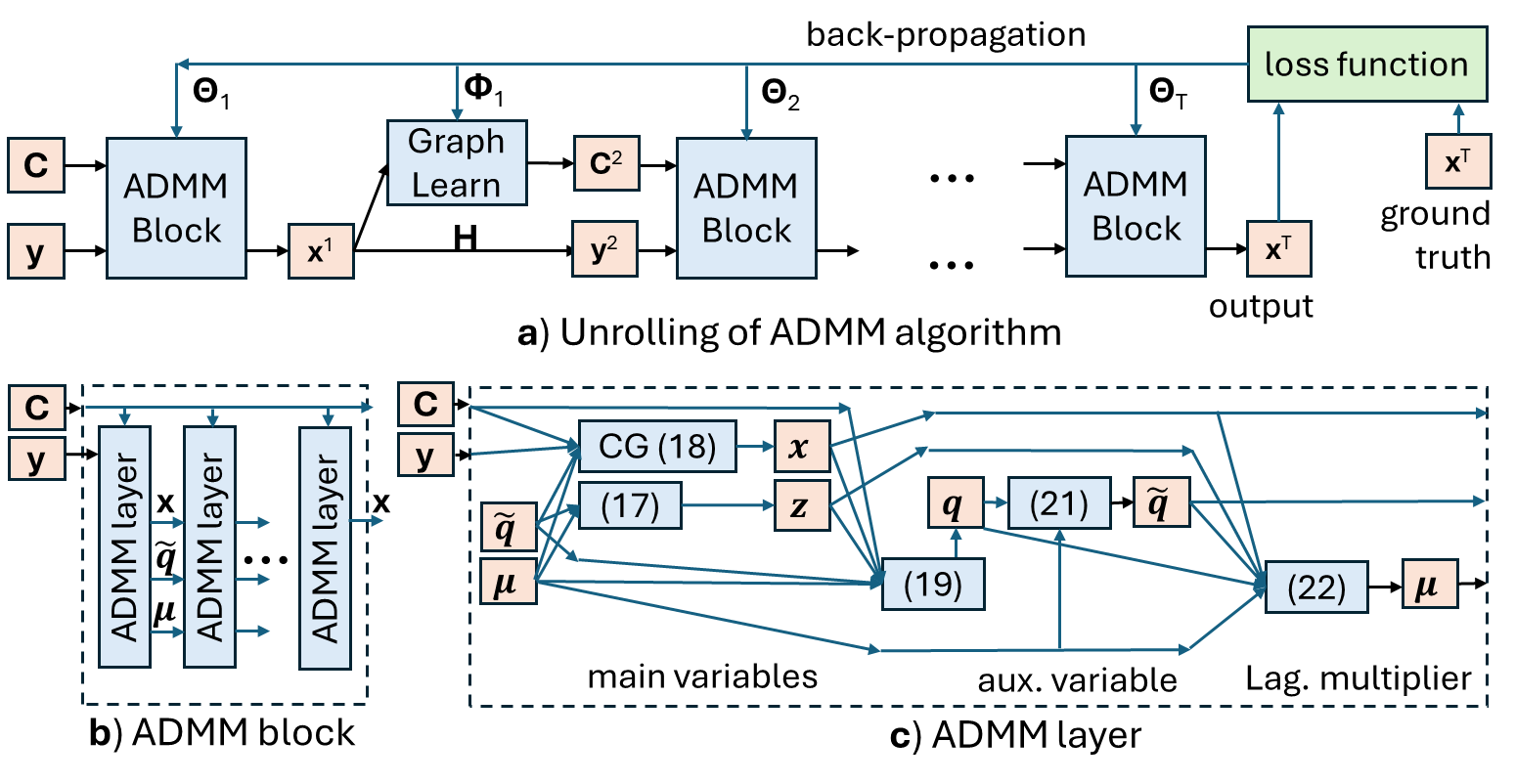}
\vspace{-0.15in}
\caption{Unrolling of GTV-based signal interpolation algorithm.  }
\label{fig:unroll_GTV}
\end{figure}

We unroll $T$ sequential pairs of an iterative interpolation algorithm (GLR- or GTV-based) with a graph learning module into an interpretable neural net. 
See Fig.\;\ref{fig:unroll_GTV}a for an illustration of the GTV-based algorithm unrolling, where the $t$-th pair of ADMM block and the graph learning module have respective parameters $\bTheta_t$ and $\bPhi_t$ that are learned from back-propagation via a defined loss function.
$\bPhi_t$ include parameters used to define feature function $F(\cdot)$ and metric matrix $\M$ in \eqref{eq:edgeWeight}, so the module can construct a graph $\cG$ specified by incidence matrix $\C^{t+1}$ given signal $\x^t$.  
In our implementation, we employ a shallow CNN to map a neighborhood of pixels centered at pixel $i$ to a low-dimensional feature $\f_i$, with a parameter size smaller than query and key matrices, $\Q$ and $\K$, in a conventional transformer. 
See Section\;\ref{subsec:exp_setup} for details.

An ADMM block contains multiple ADMM layers that are unrolled iterations of the iterative ADMM algorithm described in Section\;\ref{subsec:ADMM_opt}.
Each ADMM layer updates the main variables $\x, \z, \q$, auxiliary variable $\tilde{\q}$, and Lagrange multiplier $\bmu$ in turn using \eqref{eq:sol_ADMM1a} to \eqref{eq:sol_ADMM3}. 
ADMM weight parameter $\gamma$, as well as parameters in CG used to compute linear system \eqref{eq:sol_ADMM1b}, are learned via back-propagation. 
Specifically, two CG parameters $\alpha$ and $\beta$ that represent step size and momentum during the conjugate gradient descent step are learned.
See Appendix\;\ref{append:CG} for details.



\vspace{-0.05in}
\section{Experiments}
\label{sec:results}
\vspace{-0.05in}
\subsection{Experimental Setup}
\label{subsec:exp_setup}

\vspace{-0.05in}
All models were developed using Python 3.11. We leveraged PyTorch to implement all models and trained them using NVIDIA GeForce RTX 2080 Ti. 
To train each learned model, we used the DIV2K dataset, which contains 800 and 100 high-resolution (HR) training and validation images, respectively. 
Since the images are HR, we patchified the images into small images and used only about $1$ to $4\%$ of the patches for training and validation sets. We randomly sampled patches of $64 \times 64$ pixels to train the model. 
To test a model, we used the McM \cite{zhang2011color}, Kodak \cite{kodak1993kodak}, and Urban100 \cite{huang2015single} datasets, running each model on the whole images. 
See Appendix\;\ref{append:exp_detail} for more implementation details.

We tested model performance in two imaging applications: demosaicking and image interpolation. 
Demosaicking reconstructs a full-color image (each pixel contains RGB colors) from a Bayer-patterned image, where each pixel location has only one of Red, Green, or Blue. 
Interpolation reconstructs empty pixels missing all three colors in an image. 
To create input images, for the first application, we started from a full-color image and then removed color components per pixel according to the Bayer pattern.
For the second application, we directly down-sampled horizontally and vertically a HR image by a factor of $2$ to get the corresponding low-resolution (LR) image without any anti-aliasing filtering. 
This is equivalent to keeping every four pixels in the HR image. 

\vspace{-0.05in}
\subsection{Experimental Results}

\vspace{-0.05in}
For the first application, we evaluated our graph-based models against two variants of RSTCANet \cite{xing2022residual}, RSTCANet-B and RSTCANet-S (RST-B and RST-S for short), a SOTA framework that employs a swin transformer architecture, Menon \cite{4032820},  Malvar \cite{1326587} and bicubic interpolation. 
Menon \cite{4032820} employs a directional approach combined with an \textit{a posteriori} decision, followed by an additional refinement step. 
Malvar \cite{1326587} uses a linear filtering technique that incorporates inter-channel information across all channels for demosaicking.

The baselines for our second application are MAIN \cite{MAIN}, a multi-scale deep learning framework for image interpolation, SwinIR \cite{Liang_2021_ICCV}, and bicubic interpolation. 
SwinIR consists of three stages: shallow feature extraction, deep feature extraction, and a final reconstruction stage; see \cite{Liang_2021_ICCV} for details.
We use Peak Signal-to-Noise Ratio (PSNR) and Structural Similarity Index Measure (SSIM) \cite{wang04} as our evaluation metrics, common in image quality assessment. 


\begin{figure}
  \begin{minipage}{.45\linewidth}
    \centering    \includegraphics[height=1.5in,width=3.0in]{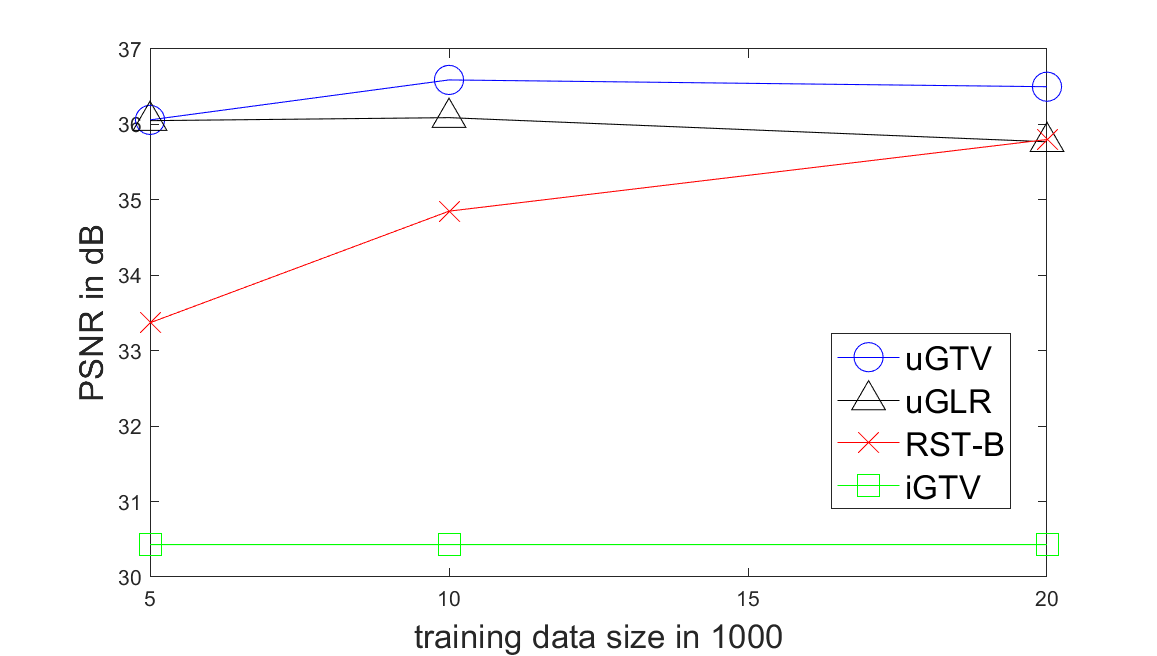}
    \vspace{-0.2in}
    \captionof{figure}{\small Demosaicking performance vs. training size for different models.}
    \label{fig:f1}
  \end{minipage}\hfill
  \begin{minipage}{.48\linewidth}
    \centering
    \begin{footnotesize}
    \begin{tabular}{c|cccc}
    \hline
    \multirow{2}{*}{Method} & \multicolumn{4}{c}{\begin{tabular}[c]{@{}c@{}}McM\\ PSNR\end{tabular}} \\
        & $\sigma=10$      & $\sigma=20$     & $\sigma=30$     & $\sigma=50$     \\
        \hline
        RST-B & 28.01 & 22.7 & 19.34 & 15.03 \\
        uGLR & 28.24 & 22.84 & 19.49 & 15.203 \\
        uGTV & \textbf{28.31} & \textbf{22.89}  & \textbf{19.56}  & \textbf{15.38}  \\ \hline
    \end{tabular}
    \end{footnotesize}
    \captionof{table}{\small Demosaicking performance in noisy scenario, where models are trained on noiseless dataset.}
    \label{table:t2}
  \end{minipage}
\end{figure}

\begin{figure}
\begin{minipage}{\linewidth}
\centering
\begin{footnotesize}
    \begin{tabular}{ccccc}
        \hline
        Method & Params\#  &\begin{tabular}{c|c}\multicolumn{2}{c}{McM}\\ PSNR &SSIM \end{tabular} & \begin{tabular}{c|c}\multicolumn{2}{c}{Kodak}\\ PSNR &SSIM \end{tabular} & \begin{tabular}{c|c}\multicolumn{2}{c}{Urban100}\\ PSNR &SSIM \end{tabular} \\
        \hline
        Bilinear &-   &\begin{tabular}{cc} 29.71 &0.9304 \end{tabular} &\begin{tabular}{cc} 28.22 &0.8898 \end{tabular} & \begin{tabular}{cc} 24.18 &0.8727 \end{tabular} \\
        RST-B \cite{xing2022residual} &931763    &\begin{tabular}{cc} 34.85 &0.9543 \end{tabular} &\begin{tabular}{cc} 38.75 &0.9857 \end{tabular} & \begin{tabular}{cc} 32.82 &0.973 \end{tabular} \\
        RST-S \cite{xing2022residual} &3162211   &\begin{tabular}{cc} 35.84 &0.961 \end{tabular} &\begin{tabular}{cc} \textbf{39.81} &\textbf{0.9876} \end{tabular} & \begin{tabular}{cc} 33.87 &0.9776 \end{tabular} \\
        Menon \cite{4032820} &-   &\begin{tabular}{cc} 32.68 &0.9305 \end{tabular} &\begin{tabular}{cc} 38.00 &0.9819 \end{tabular} & \begin{tabular}{cc} 31.87 &0.966 \end{tabular} \\
        Malvar \cite{1326587} &-   &\begin{tabular}{cc} 32.79 &0.9357 \end{tabular} &\begin{tabular}{cc} 34.17 &0.9684 \end{tabular} & \begin{tabular}{cc} 29.00 &0.9482 \end{tabular} \\
        iGLR  &-         &\begin{tabular}{cc} 29.39 &0.8954 \end{tabular} &\begin{tabular}{cc} 27.50 &0.8487 \end{tabular} & \begin{tabular}{cc} 23.13 &0.8406 \end{tabular} \\
        iGTV  &-         &\begin{tabular}{cc} 30.43 &0.8902 \end{tabular} & \begin{tabular}{cc} 28.66 &0.8422 \end{tabular} & \begin{tabular}{cc} 24.91 &0.8114 \end{tabular} \\
        uGLR  &323410    &\begin{tabular}{cc} 36.09 &0.9650 \end{tabular} & \begin{tabular}{cc} 37.88 &0.9821 \end{tabular} & \begin{tabular}{cc} 33.60 &0.9772 \end{tabular} \\
        uGTV  &323435    &\begin{tabular}{cc} \textbf{36.59} &\textbf{0.9665} \end{tabular} &\begin{tabular}{cc} 39.11 &0.9855 \end{tabular} & \begin{tabular}{cc} \textbf{34.01} &\textbf{0.9792} \end{tabular} \\ \hline
    \end{tabular}
\end{footnotesize}
\captionof{table}{ Demosaicking performance for different models, trained on 10k sample dataset.}
\label{table:t1}
\end{minipage}
\end{figure}

\begin{minipage}{\linewidth}
\centering
\begin{footnotesize}
    \begin{tabular}{ccccc}
        \hline
        Method & Params\#  &\begin{tabular}{c|c}\multicolumn{2}{c}{McM}\\ PSNR &SSIM \end{tabular} & \begin{tabular}{c|c}\multicolumn{2}{c}{Kodak}\\ PSNR &SSIM \end{tabular} & \begin{tabular}{c|c}\multicolumn{2}{c}{Urban100}\\ PSNR &SSIM \end{tabular} \\
        \hline
        Bicubic  &-        &\begin{tabular}{cc} 29.01 &0.8922 \end{tabular} &\begin{tabular}{cc} 26.75 &0.8299 \end{tabular} &\begin{tabular}{cc} 22.95 &0.7911 \end{tabular} \\
        MAIN \cite{MAIN} &10942977 &\begin{tabular}{cc} 32.72 &0.822 \end{tabular} &\begin{tabular}{cc} 28.23 &0.728 \end{tabular} &\begin{tabular}{cc} 25.46 &0.806 \end{tabular} \\
        SwinIR-lightweight \cite{Liang_2021_ICCV} &904744 &\begin{tabular}{cc} 32.24 &0.9354 \end{tabular} &\begin{tabular}{cc} 28.62 &0.8794 \end{tabular} &\begin{tabular}{cc} 25.08 &0.8553 \end{tabular} \\
        iGLR &-        &\begin{tabular}{cc} 28.53 &0.8537 \end{tabular} &\begin{tabular}{cc} 26.71 &0.8005 \end{tabular} &\begin{tabular}{cc} 22.87 &0.7549 \end{tabular} \\
        iGTV &-        &\begin{tabular}{cc} 30.41 &0.887 \end{tabular} &\begin{tabular}{cc} 28.05 &0.832 \end{tabular} &\begin{tabular}{cc} 24.26 &0.7855 \end{tabular} \\
        uGLR &319090   &\begin{tabular}{cc} 33.31 &0.9431 \end{tabular} &\begin{tabular}{cc} \textbf{29.10} &0.8870 \end{tabular} &\begin{tabular}{cc} 25.94 &0.8777 \end{tabular} \\
        uGTV &319115   &\begin{tabular}{cc} \textbf{33.36} &\textbf{0.9445} \end{tabular} &\begin{tabular}{cc} 29.08 &\textbf{0.8888} \end{tabular} &\begin{tabular}{cc} \textbf{26.12} &\textbf{0.8801} \end{tabular} \\ 
        \hline
    \end{tabular}
\end{footnotesize}
\captionof{table}{\small Interpolation performance for different models, trained on 10k sample dataset.}
\label{table:t3}
\end{minipage}

Table\;\ref{table:t1} shows the demosaicking performance for different models, where all models were trained on the same dataset and the same number of epochs ($30$), using a subset of DIV2K dataset containing $10K$ of $64 \times 64$ patches.
We observe that our unrolled GTV model (uGTV)  achieved the best overall performance, while the unrolled GLR model (uGLR) and RST-S performed similarly.
Both our models (uGTV and uGLR) performed better than RST-B while employing significantly fewer parameters.
Crucially, we observe that although our normalized edge weight $\bar{w}_{i,j}$ based on Mahalanobis distance is symmetric while the self-attention weight $a_{i,j}$ is not (due to query and key matrices $\Q$ and $\K$ not being the same in general), the directionality in the self-attention mechanism does not appear to help improve performance of the conventional transformer further, at least for image interpolation tasks. 
The iterative GTV algorithm (iGTV) without parameter optimization performed the worst, demonstrating the importance of parameter learning.  

In Fig.\;\ref{fig:f1}, we see the demosaicking performance of different models versus training data size. 
We see that for small data size, our models (uGTV and uGLR) performed significantly better than RST-B. 
This is intuitive, since a model with more parameters requires more training data in general. 
See Appendix\;\ref{append:exp_results_ap} for example visual results.

Next, we test robustness to covariate shift by testing models trained on noiseless data using a dataset artificially injected with Gaussian noise. 
Table\;\ref{table:t2} shows the demosaicking performance versus different noise variances.
We observe that our models outperformed RST-B in all noisy scenarios.




For image interpolation, we interploated a LR image to a corresponding HR image. 
We trained all models on the same dataset as the first application with the same number of epochs ($15$). 
Table\;\ref{table:t3} shows that under the same training conditions, our proposed models (uGTV and uGLR) outperformed MAIN in interpolation performance in all three benchmark datasets by about $0.7$ dB.
Note that for this application we only interpolated Y-channel from YCbCr color space for PSNR computation.
Similar to the first application, our models achieved slight performance gain while employing drastically fewer parameters; specifically, uGTV employed only about $3\%$ of the parameters in MAIN.

\vspace{-0.05in}
\section{Conclusion}
\label{sec:conclude}
\vspace{-0.05in}
By unrolling iterative algorithms that minimize one of two graph smoothness priors---$\ell_2$-norm graph Laplacian regularizer (GLR) or $\ell_1$-norm graph total variation (GTV)---we build interpretable and light-weight transformer-like neural nets for the signal interpolation problem.
The key insight is that the normalized graph learning module is akin to the self-attention mechanism in a conventional transformer architecture.
Moreover, the interpolated signal in each layer is simply the low-pass filtered output derived from the assumed graph smoothness prior, eliminating the need for the value matrix.
Experiments in two imaging applications show that interpolation results on par with SOTA can be achieved with a fraction of the parameters used in conventional transformers.

\bibliographystyle{IEEEbib}
\bibliography{ref2}

\appendix
\section{Appendix}
\label{sec:appendix}
\subsection{Full-Rankness of Matrix $\P$ in \eqref{eq:GLR_inter}}
\label{append:GLR}

Given that the underlying graph $\cG$ is positive and connected, we prove that coefficient matrix $\P$ in \eqref{eq:GLR_inter} is full-rank and thus invertible.
We prove by contradiction: suppose $\P$ is \textit{not} full-rank, and there exists a vector $\v = [\x; ~\bmu]$ such that $\P \v = \0_{N+K}$. 
Suppose we order $K$ sampled entries $\x_\cS$ before $N-K$ non-sampled entries $\x_{\bar{\cS}}$ in $\x$, \ie, $\x = [\x_\cS; \x_{\bar{\cS}}]$. 
First, given sampling matrix $\H = [\I_K ~~\0_{K,N-K}] \in \{0,1\}^{K \times N}$, focusing on the second block row of $\P$, $[\H ~~\0_{K,K}] \v = \0_K$ means $\H \x = [\I_k ~\0_{K,N-K}] [\x_\cS; \x_{\bar{\cS}}] = \0_K$.
Thus, sampled entries of $\x$ must be zeros, \ie, $\x_\cS = \0_K$.  

Second, suppose we write Laplacian $\L$ in blocks, \ie, $\L = [\L_{\cS,\cS} ~\L_{\cS,\bar{\cS}}; \L_{\bar{\cS},\cS} ~\L_{\bar{\cS},\bar{\cS}}]$. 
Then, the non-sampled rows of the first block row of $\M \v$ are
\begin{align}
\left( [2\L ~~\H^\top]
\left[ \begin{array}{c}
\x \\
\bmu \end{array} \right] \right)_{\bar{\cS}}
= 2 \left( \L_{\bar{\cS},\cS} \x_\cS + \L_{\bar{\cS},\bar{\cS}} \x_{\bar{\cS}} \right) 
\end{align}
where $(\H^\top \bmu)_{\bar{\cS}} = \0_{N-K}$ since non-sampled rows of $\H^\top = [\I_K; \0_{N-K,K}]$ are all zeros.
From above, we know $\x_{\cS} = \0_K$, and thus $([2\L ~~\H^\top] \v)_{\bar{\cS}} = \0_{N-K}$ implies that we require $\L_{\bar{\cS},\bar{\cS}} \x_{\bar{\cS}} = \0_N$.
However, since $\L_{\bar{\cS},\bar{\cS}}$ is a combinatorial Laplacian matrix for a positive sub-graph connecting non-sampled nodes \textit{plus} at least one strictly positive self-loop (representing an edge from a non-sampled node to a sample node), given $\cG$ is a connected positive graph, $\L_{\bar{\cS},\bar{\cS}}$ must be PD (see Appendix\;\ref{append:PD_L} below).
Thus, $\nexists \x_{\bar{\cS}} \neq \0$ s.t. $\L_{\bar{\cS},\bar{\cS}} \x_{\bar{\cS}} = \0_{N-K}$, a contradiction.
Therefore, we can conclude that $\P$ must be full-rank.

\subsection{Positive Definiteness of Matrix $\L_{\bar{\cS},\bar{\cS}}$ in \eqref{eq:GLR_sol2}}
\label{append:PD_L}

Given that the underlying graph $\cG$ is positive and connected, we prove that coefficient matrix $\L_{\bar{\cS},\bar{\cS}}$ in \eqref{eq:GLR_sol2} is PD.
By definition $\L \triangleq \text{diag}(\W\1_N) - \W$, where $\W$ is an adjacency matrix for a positive graph without self-loops, \ie, $W_{i,j} \geq 0, \forall i \neq j$ and $W_{i,i} = 0, \forall i$. 
Thus, $L_{i,i} = \sum_{j} W_{i,j}, \forall i$, and $L_{i,j} = -W_{i,j} \leq 0, \forall i \neq j$. 
For sub-matrix $\L_{\bar{\cS},\bar{\cS}}$, $L_{i,i} = \sum_{j \in \bar{\cS}} W_{i,j} + \sum_{j \in \cS} W_{i,j}, \forall i \in \bar{\cS}$. 
Define $\L_{\bar{\cS},\bar{\cS}}'$ as a graph Laplacian matrix for nodes in $\bar{\cS}$ considering only edges between nodes in $\bar{\cS}$, \ie, $L_{i,i} = \sum_{j \in \bar{\cS}} W_{i,j}, \forall i \in \bar{\cS}$.
Define $\D_{\bar{\cS},\bar{\cS}}'$ as a diagonal degree matrix for nodes in $\bar{\cS}$ considering only edges between $\bar{\cS}$ and $\cS$, \ie, $D'_{i,i} = \sum_{j \in \cS} W_{i,j}, \forall i \in \bar{\cS}$. 
Note that $D'_{i,i} \geq 0, \forall i$. 
We can now write $\L_{\bar{\cS},\bar{\cS}}$ as
\begin{align}
\L_{\bar{\cS},\bar{\cS}} = \L_{\bar{\cS},\bar{\cS}}' + \D'_{\bar{\cS},\bar{\cS}} .
\end{align}
$\L_{\bar{\cS},\bar{\cS}}'$ is a combinatorial graph Laplacian for a positive graph without self-loops, and thus is provably PSD \cite{cheung18}. 
$\D'_{\bar{\cS},\bar{\cS}}$ is a non-negative diagonal matrix, and thus is also PSD.
By Weyl's inequality, $\L_{\bar{\cS},\bar{\cS}}$ is also PSD. 

We prove by contradiction: suppose $\L_{\bar{\cS},\bar{\cS}}$ is not PD, and $\exists \x \neq \0$ such that $\x^\top \L_{\bar{\cS},\bar{\cS}} \x = 0$. 
$\x^\top \L_{\bar{\cS},\bar{\cS}} \x = 0$ iff $\x^\top \L_{\bar{\cS},\bar{\cS}}' \x = 0$ and $\x^\top \D'_{\bar{\cS},\bar{\cS}} \x = 0$ simultaneously. 
Denote by $\bar{\cS}_1$ and $\bar{\cS}_2$ the indices of nodes in $\bar{\cS}$ with and without connections to nodes in $\cS$, respectively. $\bar{\cS}_1 \neq \emptyset$, since $\cG$ is connected. 
Suppose first $\bar{\cS}_2 = \emptyset$.
Then $\D'_{\bar{\cS},\bar{\cS}}$ has strictly positive diagonal entries and is PD, and there is no $\x \neq \0$ s.t. $\x^\top \D_{\bar{\cS},\bar{\cS}}' \x = \0$, a contradiction.

Suppose now $\bar{\cS}_2 \neq \emptyset$.
First, $\x^\top \D'_{\bar{\cS},\bar{\cS}} \x = 0$ implies $\x_{\bar{\cS}_1} = \0$.
Then,
\begin{align}
\x^\top \L_{\bar{\cS},\bar{\cS}}' \x = \sum_{i,j \in \bar{\cS}} W_{i,j} (x_i - x_j)^2 \geq \sum_{i \in \bar{\cS}_1, j \in \bar{\cS}_2} W_{i,j} (x_i - x_j)^2 .
\end{align}
Since each term in the sum is non-negative, the sum is zero only if for each $(i,j) \in \cE$ where $i \in \bar{\cS}_1$ and $j \in \bar{\cS}_2$, $0 = x_i = x_j$. 
For nodes $k \in \bar{\cS}_2$ connected only to nodes $j \in \bar{\cS}_2$ connected to $i \in \bar{\cS}_1$, $x_k = x_j = x_i = 0$ necessarily, and for nodes $l \in \bar{\cS}_2$ connected to $k \in \bar{\cS}_2$ must have $x_l = x_k = 0$, and so on.
Thus, $\x = \0$, a contradiction.
Thus, we can conclude that $\L_{\bar{\cS},\bar{\cS}}$ is PD.

\subsection{Derivation of \eqref{eq:sol_ADMM1a}, \eqref{eq:sol_ADMM1b} and \eqref{eq:sol_ADMM1c}}
\label{append:sol_ADMM1}

We define $\bphi = [\z; \x; \q_1; \q_2]$ and rewrite the objective \eqref{eq:obj_ADMM1} to
\begin{align}
\min_{\bphi} ~& \left[ \begin{array}{c}
\1_M \\
\0_{N+2M}
\end{array} \right]^\top \bphi + (\bmu^t)^\top \left( \left[\begin{array}{c}
\A \\
\0_{2M,M+N} ~ \I_{2M} 
\end{array} \right] \bphi - \left[ \begin{array}{c}
\b \\
\tilde{\q}^t
\end{array} \right] \right) 
\nonumber \\
&+ \frac{\gamma}{2} \left\| 
\underbrace{\left[ \begin{array}{c}
\A \\
\0_{2M,M+N} ~ \I_{2M} \end{array} \right]}_{\B}  \bphi - \left[ \begin{array}{c}
\b \\
\tilde{\q}^t
\end{array} \right]
\right\|^2_2 .
\label{eq:obj_ADMM1_append}
\end{align}
\eqref{eq:obj_ADMM1_append} is a convex quadratic objective, and so we take the derivative w.r.t. $\bphi$ and set it to $\0$:
\begin{align}
& \left[ \begin{array}{c}
\1_M \\
\0_{N+2M}
\end{array} \right] + 
\left[ \A^\top \begin{array}{c}
\0_{M+N,2M} \\
\I_{2M} \end{array} \right] (\bmu^t)
\nonumber \\
&+ \frac{\gamma}{2} \left(
2 \left[ \A^\top \begin{array}{c}
\0_{M+N,2M} \\
\I_{2M} \end{array} \right]
\left[\begin{array}{c}
\A \\
\0_{2M,M+N} ~ \I_{2M} 
\end{array} \right] \bphi
- 2 \left[ \A^\top \begin{array}{c}
\0_{M+N,2M} \\
\I_{2M} \end{array} \right] \left[ \begin{array}{c}
\b \\
\tilde{\q}^t
\end{array} \right] \right) = \0 .
\label{eq:derivative0}
\end{align}

Given that $\B$ is the following matrix:
\begin{align}
    \B = \left[ \begin{array}{cccc}
        \I_M     &-\C     &-\I_M     &\0_M        \\
        \I_M     &\C      &\0_M      &-\I_M    \\
        \0_{K,M} &\H       &\0_{K,M}  &\0_{K,M} \\
        \0_{M,M} &\0_{M,N} &\I_M      &\0_M        \\
        \0_{M,M} &\0_{M,N} &\0_M      &\I_M     
    \end{array} \right] 
\end{align}
Hence, $\B^\top \B$ is
\begin{align}
    \B^\top\B = \left[ \begin{array}{cccc}
        2\I_M &\0_{M,N} &-\I_M  &-\I_M \\
        \0_{N,M} &2\C^\top \C + \H^\top \H &\C^\top &-\C^\top \\
        -\I_M  &\C &2\I_M &\0_{M,M} \\
        -\I_M  &-\C &\0_{M,M} &2\I_M \\
    \end{array} \right] .
\end{align}
Note that adding two of row $1$ to rows $3$ and $4$, we get $[2 \I_M \; \0_{M,N} \; \0_{M,M} \; \0_{M,M}]$.

Solving for $\bphi$ in \eqref{eq:derivative0}, we get 
\begin{align}
    \gamma \B^\top\B\bphi &= - \left[ \begin{array}{c}
\1_M \\
\0_N \\
\0_M \\
\0_M \\
\end{array} \right] - \B^\top \left( \left[ \begin{array}{c}
\bmu_a^t\\
\bmu_b^t\\
\bmu_c^t\\
\bmu_d^t\\
\bmu_e^t
\end{array} \right] - \gamma \left[ \begin{array}{c}
\0_M\\
\0_M\\
\y \\
\tilde{\q}_1^t \\
\tilde{\q}_2^t 
\end{array} \right] \right) \\
&= 
\left[ \begin{array}{c}
-\1_M - \bmu_a^t - \bmu_b^t \\
\C^\top \bmu_a^t - \C^\top \bmu_b^t - \H^\top \bmu_c^t + \gamma \H^\top \y \\
\bmu_a^t - \bmu_d^t + \gamma \tilde{\q}_1^t \\
\bmu_b^t - \bmu_e^t + \gamma \tilde{\q}_2^t
\end{array} \right]
\label{eq:linSys_general}
\end{align}
We can solve for $\z^{t+1}$ directly; by adding two of row $1$ to rows $3$ and $4$ of \eqref{eq:linSys_general}, we get
\begin{align}
2 \gamma \z^{t+1} &= -2 (\1_M) - \bmu_a^t - \bmu_b^t - \bmu_d^t - \bmu_e^t + \gamma (\tilde{\q}_1^t + \tilde{\q}_2^t) 
\nonumber \\
\z^{t+1} &= \red{-}\frac{1}{\gamma} \1_M - \frac{1}{2\gamma} \left( \bmu_a^t + \bmu_b^t + \bmu_d^t + \bmu_e^t \right) + \frac{1}{2} (\tilde{\q}_1^t + \tilde{\q}_2^t) .
\end{align}

Subtracting row $4$ from row $3$ of \eqref{eq:linSys_general}, we get
\begin{align}
2\gamma \C \x^{t+1} + 2\gamma (\q_1^{t+1} - \q_2^{t+1}) &= \bmu_a^t - \bmu_b^t - \bmu_d^t + \bmu_e^t + \gamma (\tilde{\q}_1^t - \tilde{\q}_2^t)
\nonumber \\
\gamma(\q_1^{t+1} - \q_2^{t+1}) &= - \gamma \C \x^{t+1} + \frac{1}{2} \left( \bmu_a^t - \bmu_b^t - \bmu_d^t + \bmu_e^t \right) + \frac{\gamma}{2} (\tilde{\q}_1^t - \tilde{\q}_2^t) .
\end{align}
Thus, row 2 can be rewritten as
\begin{align}
\gamma (2 \C^\top \C + \H^\top \H) \x^{t+1} + \C^\top \gamma ( \q_1^{t+1} - \q_2^{t+1} ) &= \C^\top \bmu_a^t - \C^\top \bmu_b^t - \H^\top \bmu_c^t + \gamma \H^\top \y
\nonumber \\
\gamma (\C^\top \C + \H^\top \H) \x^{t+1} &= 
 \frac{1}{2} \C^\top \left( \bmu_a^t - \bmu_b^t + \bmu_d^t - \bmu_e^t \right) - \H^\top \bmu_c^t 
\nonumber \\
& ~~ - \frac{\gamma}{2} \C^\top (\tilde{\q}_1^t - \tilde{\q}_2^t) + \gamma \H^\top \y .
\end{align}
Finally, from rows $3$ and $4$, $\q_1^t$ and $\q_2^t$ can be computed as
\begin{align}
\q_1^t &= \frac{1}{2} \left( \z^{t+1} - \C \x^{t+1} \right) + \frac{1}{2\gamma}(\bmu_a^t - \bmu_d^t + \gamma \tilde{\q}_1^t)
\nonumber \\
\q_2^t &= \frac{1}{2} \left( \z^{t+1} + \C \x^{t+1} \right) + \frac{1}{2\gamma}(\bmu_b^t - \bmu_e^t + \gamma \tilde{\q}_2^t) .
\end{align}


\subsection{Invertibility of \texorpdfstring{$\boldsymbol{\cL} = \C^\top \C + \H^\top \H$}{L}}
\label{append:invertibility}

Clearly $\boldsymbol{\cL} = \C^\top \C + \H^\top \H$ is real, symmetric, and positive semi-definite.  Thus its eigenvalues are real and non-negative.  To show that it is invertible, it suffices to show that its minimum eigenvalue $\lambda_{\min}(\boldsymbol{\cL})$ is strictly greater than zero.  But $\lambda_{\min}(\boldsymbol{\cL}) = \min_{\x:||\x||=1} \x^\top \cL \x$. Hence it suffices to show that $\x^\top \boldsymbol{\cL} \x=0$ implies $\x=\0$.  Now observe that $\x^\top \boldsymbol{\cL} \x = \sum_{(i,j)\in\cE} w_{i,j}^2(x_i - x_j)^2 + \sum_{i\in\cS} x_i^2$, where $\cS$ is the set of nodes with constrained values.  If $\x^\top \boldsymbol{\cL} \x=0$ then all terms must be zero, meaning that all $x_i$ in $\cS$ are zero, and hence all of their neighbors, and all of {\em their} neighbors, and so on.  Thus, $\boldsymbol{\cL}$ is invertible if there exists at least one node with a constrained value (i.e., a self-loop) in each connected component of the graph.

\subsection{Derivation of \eqref{eq:sol_ADMM2}}
\label{append:sol_ADMM2}

We derive the solution to optimization \eqref{eq:obj_ADMM2}.
Ignoring the first convex but non-smooth term $g(\tilde{\mathbf{q}})$, the remaining two terms in the objective are convex and smooth.
Taking the derivative w.r.t. variable $\tilde{\mathbf{q}}$ and setting it to $\mathbf{0}$, we get
\begin{align}
-\bmu^t_{d} - \gamma \mathbf{q}^{t+1} + \gamma \tilde{\mathbf{q}}^* &= \mathbf{0}_{M}
\nonumber \\
\tilde{\mathbf{q}}^* &= \mathbf{q}^{t+1} + \frac{1}{\gamma} \boldsymbol{\mu}^t_{d} .
\label{eq:appendThres}
\end{align}
This solution is valid iff $g(\tilde{\mathbf{q}}^*) = 0$; otherwise the first term $g(\tilde{\mathbf{q}}^*)$ dominates and $\tilde{\mathbf{q}}^* = \mathbf{0}_{M}$.
Given that \eqref{eq:appendThres} can be computed entry-by-entry separately, \eqref{eq:sol_ADMM2} follows.



\subsection{Example of Edge Weight Normalization for the Incidence Matrix}
\label{append:normalize}

\begin{figure}[tb]
\centering
\begin{subfigure}{.47\textwidth}
\centering
\includegraphics[width = 0.70\linewidth]{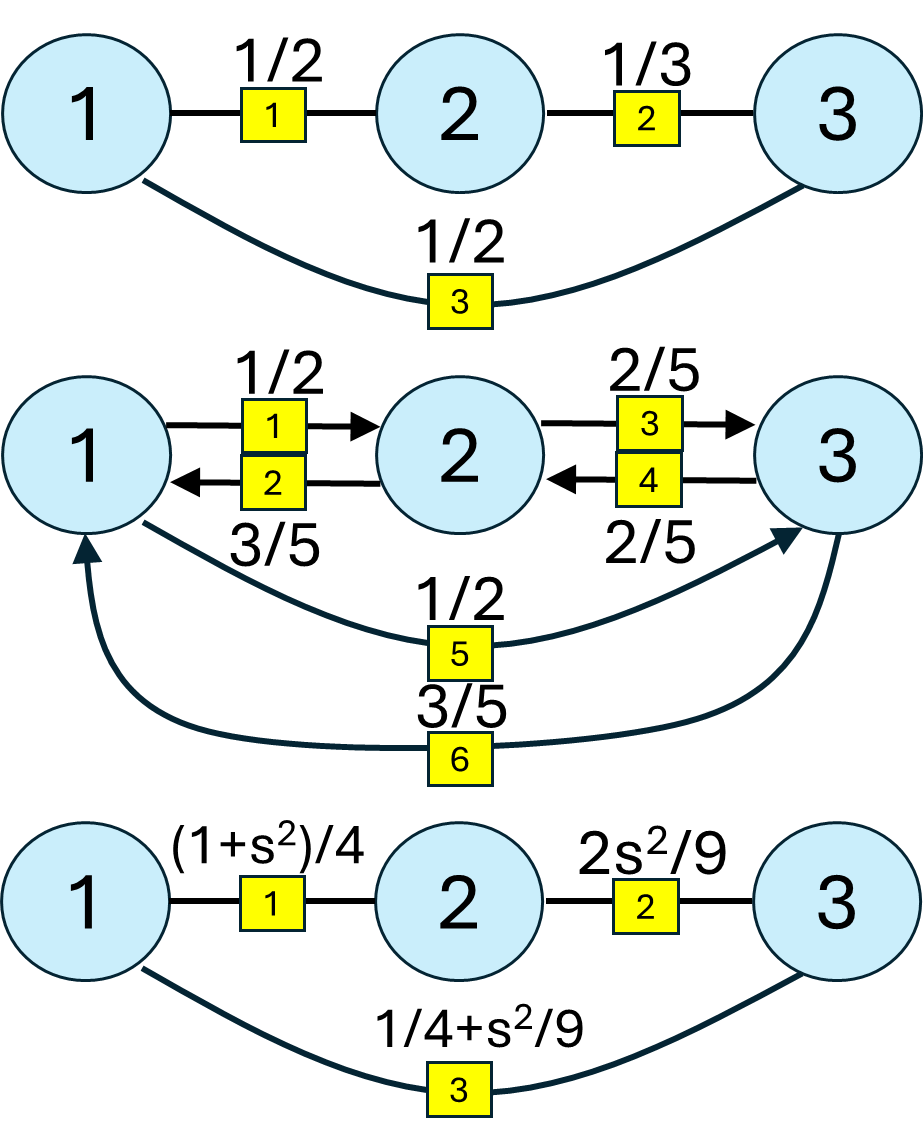}
\end{subfigure}
\begin{subfigure}{.48\textwidth}
\centering
\begin{align}
\C &= \left[\begin{array}{ccc}
1/2 & -1/2 & 0 \\
0 & 1/3 & -1/3 \\
1/2 & 0 & -1/3
\end{array}\right] 
\nonumber 
\end{align}
\begin{align}
\bar{\C} &= \left[\begin{array}{ccc}
1/2 & -1/2 & 0 \\
-3/5 & 3/5 & 0 \\
0 & 2/5 & -2/5 \\
0 & -2/5 & 2/5 \\
1/2 & 0 & -1/2 \\
-3/5 & 0 & 3/5 
\end{array}\right]
\nonumber 
\end{align}
\begin{align}
\bar{\L} &= \left[\begin{array}{ccc}
\frac{1}{2}+\frac{s^2}{4}+\frac{s^2}{9} & -\frac{1+s^2}{4} & -\frac{1}{4}-\frac{s^2}{9} \\
-\frac{1+s^2}{4} & \frac{1+s^2}{4} + \frac{2s^2}{9} & - \frac{2s^2}{9} \\
-\frac{1}{4}-\frac{s^2}{9} & - \frac{2s^2}{9} & \frac{1}{4} + \frac{s^2}{3}
\end{array}\right]
\nonumber 
\end{align}
\end{subfigure}
\caption{3-node graph for incidence matrix $\C$ (top), 3-node graph for normalized incidence matrix $\bar{\C}$ (middle), 3-node graph for graph Laplacian $\bar{\L} = \bar{\C}^\top \bar{\C}$ where $s=\textstyle \frac{6}{5}$ (bottom).  }
\label{fig:incidence_ex}
\end{figure}

In Fig.\;\ref{fig:incidence_ex}(top), we show an example three-node \textit{undirected} graph with three edge weights $w_{1,2} = 1/2$, $w_{1,3} = 1/2$, and $w_{2,3} = 1/3$, and the corresponding incidence matrix $\C$. 
Normalizing edge weights using \eqref{eq:rwEdgeWeight}, we see in Fig.\;\ref{fig:incidence_ex}(middle) a \textit{directed} graph with six edges where the sum of normalized edge weights leaving a node is $1$, resulting in normalized incidence matrix $\bar{\C}$. 
Finally, we see in Fig.\;\ref{fig:incidence_ex}(bottom) an \textit{undirected} graph with three edges corresponding to graph Laplacian $\bar{\L} = \bar{\C}^\top \bar{\C}$. 
Note that $\|\bar{\C} \1_3\|_1 = \0_6$ as expected.

\subsection{Parameters Learning in Conjugate Gradient Algorithm (CG)}
\label{append:CG}

The linear systems that we need to solve---\eqref{eq:GLR_sol2} for GLR minimization and \eqref{eq:sol_ADMM1b} for GTV minimization---have the follow form,
\begin{align}
\boldsymbol{\cL}\x = \b .
\label{eq:linear_system}
\end{align}
Given $\boldsymbol{\cL}$ is PD, we consider the minimization problem,
\begin{equation}
    \min_{\x} Q(\x) = \frac{1}{2} \x^\top \boldsymbol{\cL} \x - \b^\top\x
\end{equation}
with gradient
\begin{equation}
\frac{\delta Q(\x)}{ \delta \x } = \boldsymbol{\cL} \x - \b .
\end{equation}
Thus, the simple gradient descent has the following update rules with $\alpha_t$ commonly known as learning rate,
\begin{align}
    \g^t &= \boldsymbol{\cL} \x^{t} - \b = \g^{t-1} - \alpha_t \boldsymbol{\cL} \g^{t-1} \\
    \x^{t+1} &= \x^{t} - \alpha_t \g^t .
\end{align}
Next, a momentum term $\beta_t$ and the cumulative gradients term $\v^t$ are added, resulting in the well-known Accelerated Gradient Descent Algorithm \red{\cite{nesterov2013introductory}}, also known as Conjugated Gradient Descent. 
The new update rules are
\begin{align}
    \g^t &= \g^{t-1} - \alpha_t \boldsymbol{\cL} \v^{t-1} \\
    \v^{t} &= \g^t + \beta_t \v^{t-1} \\
    \x^{t+1}   &= \x^t - \alpha_t \v^{t}
\end{align}
where both $\alpha_t$ and $\beta_t$ in each iteration $t$ are considered trainable parameters.


\subsection{Experimental Details}
\label{append:exp_detail}

Initially, we developed our \textit{unrolled} ADMM model without training any parameters. The target signal $\x$ was estimated using linear interpolation of known values in $
5 \times 5$ pixel neighborhood. RGB values and pixel locations were combined to form feature vectors for computing edge weights $w_{i,j}$, which were shared across the three channels.
The metric matrix $\M$ was initialized as a diagonal matrix with all entries set to $1.5$. Vectors $\mu_a$, $\mu_b$, $\mu_c$, $\mu_d$, and $\mu_e$ were initialized with all entries equal to $0.1$. The parameters $\gamma$, $\alpha$, and $\beta$ in CG were set to $10$, $0.5$, and $0.3$, respectively.
For the \textit{learned} ADMM block, the parameters $\gamma$, $\alpha$, $\beta$, and the metric matrix $\M$ were learned during training. A training dataset was created consisting of 5000, 10000, or 20000 image patches, each of size $64 \times 64$, to train the model.


All matrix multiplications in our models are implemented to take advantage of the sparsity of the constructed graphs, which were restricted to be a window of size $5\times 5=25$ nearest neighbors of each pixel. 
This ensures that our graphs are always \textit{connected} and \textit{sparse}. 
We stacked vertically 4 Graph Learning modules coupled with ADMM block, so that we can learn multiple graphs in parallel.
This is commonly known as multi-head in the transformer architecture. 
We also stacked $5$ graph learning modules and ADMM blocks horizontally to further learn more graphs. 
In all ADMM blocks, we set the number of ADMM iterations to $5$ and the number of CG iterations to $10$.


To extract high-level features, a shallow CNN was employed, consisting of four convolutional layers with 48 feature maps (12 features for each graph learning head). After each convolutional layer, a ReLU activation function was applied. Convolutional layers utilized $3 \times 3$ kernels without any down-sampling, generating 48 features for each pixel.
The feature maps were divided into four sets, with each set serving as the input for a Graph Learning module, allowing for the parallel production of four graphs. A simple weighted average scheme was then used to combine the outputs into a single output $\x^t$. Additionally, skip connections were introduced between the convolutional layers to facilitate the training process. 

\subsection{Additional Experimental Results}
\label{append:exp_results_ap}

\begin{figure}
\centering
\sbox{\bigpicturebox}{%
  \subfloat[Urban100: image062.png\label{subfig-1:dummy}]{%
       \includegraphics[width=0.5\textwidth]{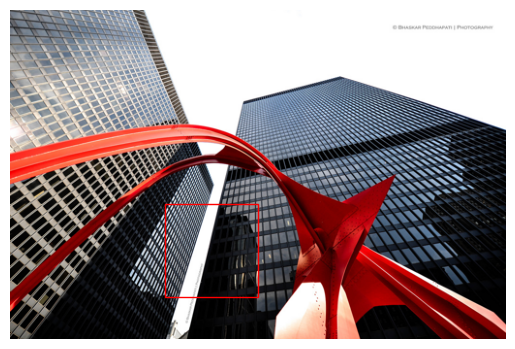}
     }%
}%
\usebox{\bigpicturebox}\hfill
\begin{minipage}[b][\ht\bigpicturebox][b]{.45\textwidth}
\subfloat[Original\label{subfig-2:dummy}]{%
       \includegraphics[width=0.30\textwidth]{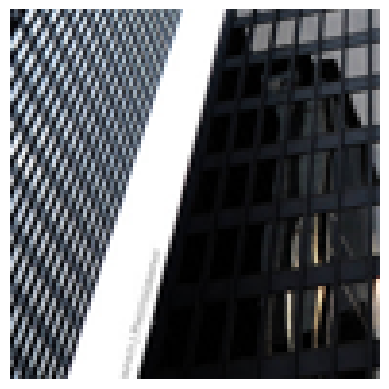}
     }%
     \hfill
     \subfloat[iGTV\label{subfig-4:dummy}]{%
       \includegraphics[width=0.30\textwidth]{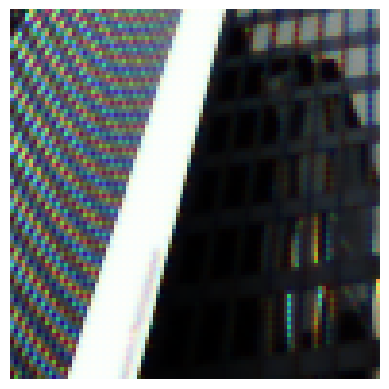}
     }%
     \hfill
     \subfloat[uGTV\label{subfig-3:dummy}]{%
       \includegraphics[width=0.30\textwidth]{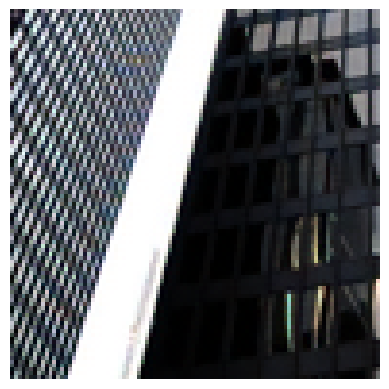}
     }%
\vfill
     \subfloat[uGLR\label{subfig-4:dummy}]{%
       \includegraphics[width=0.30\textwidth]{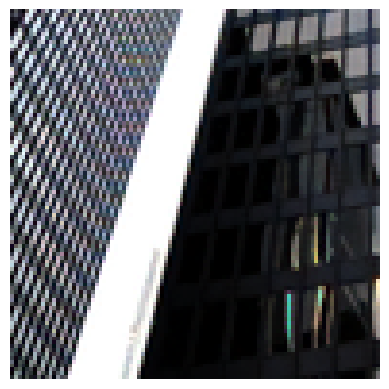}
     }%
\subfloat[RST-B\label{subfig-4:dummy}]{%
       \includegraphics[width=0.30\textwidth]{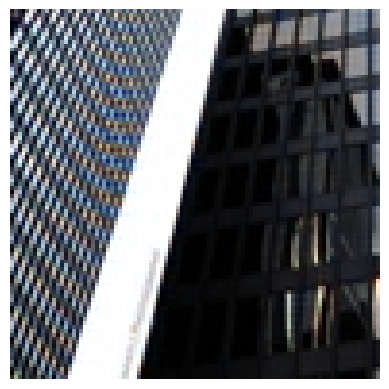}
     }%
     \hfill
     \subfloat[RST-S\label{subfig-4:dummy}]{%
       \includegraphics[width=0.30\textwidth]{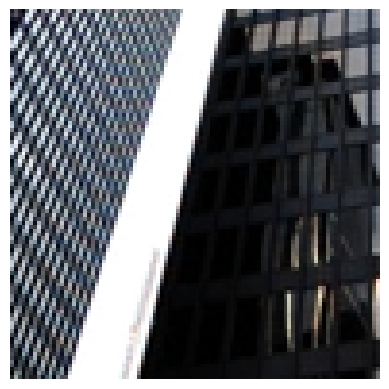}
     }%
     \vfill
\end{minipage}
\caption{Visual demosaicking results for image \textit{Urban062}.}
\label{fig:visfordem}
\end{figure}

Fig.\;\ref{fig:visfordem} shows visual results for a test image for all models, including our uGTV, uGLR and the baselines. Two variants of RSTCANet, uGTV and uGLR are trained on $10000$ images patches of size $64\times 64$ for $30$ epochs. 
We observe that our two models, especially uGTV, has better performance compared to RST-B and comparable performance with RST-S in selected high-frequency area.

\end{document}